\documentclass{article}

\usepackage[nonatbib]{neurips_2023}

\usepackage[utf8]{inputenc} 
\usepackage[T1]{fontenc}    
\usepackage{hyperref}       
\usepackage{url}            
\usepackage{booktabs}       
\usepackage{amsfonts}       
\usepackage{nicefrac}       
\usepackage{microtype}      
\usepackage{xcolor}         
\usepackage{bm}
\usepackage{amsmath}
\usepackage{graphicx}
\usepackage[capitalize]{cleveref}
\usepackage{subfig}
\usepackage{wrapfig} 
\usepackage{makecell}
\usepackage{multirow}

\title{Edit-DiffNeRF: Editing 3D Neural Radiance Fields using 2D Diffusion Model}

\author{Lu Yu\\
James Cook University\\
	{\tt\small lu.yu@my.jcu.edu.au}
	\and
	Wei Xiang\thanks{Corresponding author.}\\
	La Trobe University\\
	{\tt\small w.xiang@latrobe.edu.au}
  \and
	Kang Han\\
	La Trobe University\\
	{\tt\small k.han@latrobe.edu.au}
}

\begin{document}

\maketitle

\begin{abstract}
Recent research has demonstrated that the combination of pretrained diffusion models with neural radiance fields (NeRFs) has emerged as a promising approach for text-to-3D generation. Simply coupling NeRF with diffusion models will result in cross-view inconsistency and degradation of stylized view syntheses. To address this challenge, we propose the Edit-DiffNeRF framework, which is composed of a frozen diffusion model, a proposed delta module to edit the latent semantic space of the diffusion model, and a NeRF. Instead of training the entire diffusion for each scene, our method focuses on editing the latent semantic space in frozen pretrained diffusion models by the delta module. This fundamental change to the standard diffusion framework enables us to make fine-grained modifications to the rendered views and effectively consolidate these instructions in a 3D scene via NeRF training. As a result, we are able to produce an edited 3D scene that faithfully aligns to input text instructions. Furthermore, to ensure semantic consistency across different viewpoints, we propose a novel multi-view semantic consistency loss that extracts a latent semantic embedding from the input view as a prior, and aim to reconstruct it in different views. Our proposed method has been shown to effectively edit real-world 3D scenes, resulting in 25\% improvement in the alignment of the performed 3D edits with text instructions compared to prior work.
\end{abstract}

\section{Introduction}
With the growing prevalence of efficient 3D reconstruction techniques like Neural Radiance Fields (NeRFs), generating photo-realistic synthetic views of real-world 3D scenes has become increasingly feasible \cite{MildenhallSTBRN20}. Meanwhile, the demand for manipulating 3D scenes is rapidly increasing due to the broad range of content re-creation use cases \cite{WangCH0022}. A number of recent works have introduced extensions for editing NeRFs by e.g., operating on explicit 3D representations \cite{YuanSLMJ022}, training models to enable color modifications or the removal of certain objects \cite{LiuZZ0ZR21}, or utilizing diffusion models to perform text-to-3D synthesis \cite{poole2022dreamfusion}. These advancements have expanded the versatility of implicit volumetric representations. However, they are currently limited to changing the color and shape, or require complex and computationally expensive training procedures for each editing task. 

Editing NeRFs is a crucial and challenging task. As NeRF is an implicit representation optimized per scene, the standard tools used for editing explicit representations cannot be directly employed \cite{UyHSBG20, Wang0LFY0XJ21}. Moreover, in contrast to 2D image editing, where a single image input is able to provide sufficient guidance for manipulation, NeRFs require multi-view information to guide their content modification \cite{seo2023let}. EditingNeRF \cite{LiuZZ0ZR21} was the first attempt to enable editing on the implicit radiance field. To learn the geometry and appearance of various models, they incorporate the shape code and color code for better disentanglement. Nevertheless, EditingNeRF is limited to shape and color manipulation, or the removal of local parts of an object. Yang et al. \cite{Yang0XLZB0C21} introduce a technique that combines NeRF with explicit 3D geometry representations, to enable more efficient and accurate reconstruction of 3D objects. Yuan et al. \cite{YuanSLMJ022} further improve the performance by enabling users to perform user-controlled shape deformation in a 3D scene. However, their rendered results still suffer from significant shortcomings in terms of photorealism and geometric accuracy.  

Recently, diffusion models have been proposed as a promising approach in incorporating NeRF-like models to produce multi-view consistent images \cite{RombachBLEO22}. This is done either by optimizing NeRFs via a pretrained diffusion model such as DreamFusion \cite{poole2022dreamfusion}, or via an unconditional generative diffusion model that can be trained with 2D images \cite{Karnewar16509}. Although these approaches can generate 3D models from any given text prompts, they currently lack fine-grained control over the synthesized views. More importantly, they cannot be directly used to edit real-captured NeRFs of fully observed 3D scenes. Haque et al. \cite{haque2023instructnerf2nerf} propose Instruct-NeRF2NeRF, which extracts shape and appearance features from a pretrained 2D diffusion model (i.e., InstructPix2Pix \cite{Brooks09800}) to gradually edit the rendered images while optimizing the underlying 3D scene. This method, however, shares several limitations with InstructPix2Pix \cite{Brooks09800}, such as significant multi-view inconsistencies and notable rendering artifacts including noise and blurring.

\textbf{Our method.} To solve the aforementioned problems and perform high-fidelity text-to-3D editing with pretrained NeRFs, we propose a framework dubbed Edit-DiffNeRF to edit the semantic latent space within frozen pretrained diffusion models from an input prompt. Which enables us to make fine-grained modifications to the rendered views of NeRF, offering enhanced control and accuracy in the optimization process of an underlying scene. Our Edit-DiffNeRF is composed of a frozen diffusion model, a proposed delta module to edit the latent space of a diffusion model, and a NeRF. Our proposed framework for text-to-3D editing involves two key steps. First, in order to perform the editing for the rendered views, we utilize the diffusion prior to generate a latent semantic embedding for each view and then apply our proposed delta module to edit the embedding. This delta module is optimized using a CLIP distance loss function. After training, it is able to produce edited images based on the input text instruction. Second, we proceed to train the NeRF using those edited images, leveraging the modifications made through the delta module. In order to ensure the multi-view consistency of a 3D scene, we propose a multi-view semantic consistency loss to reconstruct consistent latent features in the latent space from different views. To evaluate its effectiveness, we evaluate our Edit-DiffNeRF on a variety of real-captured NeRF scenes published by \cite{haque2023instructnerf2nerf}. Extensive experimental results demonstrate 25\% improvement in the alignment of the performed 3D edits with the text instructions compared to Instruct-NeRF2NeRF \cite{haque2023instructnerf2nerf}.
 
\section{Related work}
\subsection{Editing NeRFs}
NeRFs \cite{MildenhallSTBRN20} have emerged as a prominent approach for rendering photo-realistic views from multiple views of 3D scenes. Recently, there has been ongoing research focusing on advancing the capabilities of editing NeRFs. One strategy is to integrate physics-based inductive biases into the training procedure of NeRFs, thereby enabling effective modifications in materials or scene lighting \cite{SrinivasanDZTMB21, VerbinHMZBS22, MunkbergCHES0GF22}. Alternatively, different approaches have been proposed involving the utilization of bounding boxes and conditions encoded on NeRFs, to allow easy manipulations of different objects \cite{OstMTKH21, YuG022}. For instance, EditingNeRF \cite{LiuZZ0ZR21} was the first attempt to enable editing on the implicit radiance field, where the 3D scene is conditioned via a shape code and an appearance code. This approach enables modifications to be made to the shape of objects in the scene as well as the colors. NeRF-Editing \cite{YuanSLMJ022} is another work for performing geometry editing on 3D scenes, which provides an intuitive interface for making fine-grained modifications to the 3D shape and structure of NeRF scenes. A recent work, Instruct-NeRF2NeRF \cite{haque2023instructnerf2nerf}, proposes a method for editing NeRF scenes with text prompts based on a pretrained diffusion framework InstructPix2Pix \cite{Brooks09800}. However, it employs a diffusion model that operates on a single view at a time, which can lead to similar artifacts encountered in InstructPix2Pix, such as the presence of duplicated or distorted objects. In this work, we instead focus on enabling the manipulation of the latent semantic space of a pretrained diffusion model, thereby allowing the editing of rendered views of NeRF and facilitating the optimization of a 3D NeRF scene.

\subsection{3D content generation}
The recent advancements in pretrained large-scale models have significantly accelerated progress in the field of generating 3D content from scratch, allowing for the fast and effective creation of 3D scenes. Some works optimize NeRFs by vision-language models such as CLIP \cite{KhalidXBP22}. CLIP-NeRF \cite{WangCH0022} introduces a disentangled conditional NeRF that allows individual control over both shape and appearance via a learned deformation field. Dream Fields \cite{JainMBAP22} combines the powerful CLIP with an optimization-based process to train NeRFs. CLIP has also been used in the recent state-of-the-art model DreamFusion \cite{poole2022dreamfusion} with a loss derived from the distillation of a 2D diffusion model to render high-fidelity coherent 3D objects. While these approaches are able to generate 3D scenes based on arbitrary text inputs, they often fall short of generalizing to real-world scenes. In parallel, recent works like SparseFusion \cite{Zhou00792} have delved into the concept of grounding by utilizing one or a few input views, effectively hallucinating the unseen parts of a scene. However, all of these generative models encounter the same challenge, i.e., consolidating the diverse and inconsistent outputs of a 2D diffusion model into a unified and consistent 3D scene. In our work, we aim to edit a real-captured 3D scene by an edited latent semantic embedding. Furthermore, we propose a novel multi-view semantic consistency loss to ensure 3D consistency across different viewpoints of a scene.

\section{Preliminaries}
\subsection{Denoising diffusion probabilistic models (DDPMs)}
\label{sec3.1}
Given a set of training views $\left\{\bm{x}^i \right\}_{i=1}^N \in \mathcal{I}$ for a 3D scene, The goal of generative models, e.g., Denoising Diffusion Probabilistic Models (DDPMs) \cite{HoJA20} is to optimize the parameters $\theta$ of a model that closely approximates the data distribution $p(\bm{x})$. DDPM proposes to learn the data distribution by gradually transforming a sample from a tractable noise distribution toward a target distribution. Diffusion models typically include a deterministic forward process $q(\bm{x}_t|\bm{x}_{t-1})$ that gradually adds noise to the sample such that:
\begin{equation}
  q(\bm{x}_t|\bm{x}_{t-1}) := \mathcal{N}(\bm{x}_t; \sqrt{1-\beta_t} \bm{x}_{t-1}, \beta_t \bm{I}), \quad t \in (0, T],
\end{equation}
where $\beta_t$ is the $t$-th variance schedule. The model then learns the reverse (denoising) process with a neural network $\mathcal{D}_{\theta}(\bm{x}_t, t)$, which performs denoising steps by progressively removing noise and predicts $\hat{\bm{x}}_0$ from $\bm{x}_t$. The denoising process similarly uses a Gaussian distribution:
\begin{equation}
  p_{\theta}(\bm{x}_{t-1}|\bm{x}_t) := \mathcal{N}(\bm{x}_{t-1};\mu_{\theta}(\bm{x}_t,t), \Sigma_{\theta}(\bm{x}_t,t)),
\end{equation}
where $\mu_{\theta}$ and $\Sigma_{\theta}$ are the mean and variance, respectively.

\subsection{Latent diffusion models}
Latent diffusion models \cite{RombachBLEO22} obtain efficiency improvements compared to DDPMs \cite{HoJA20} by leveraging the latent space of a pretrained variational autoencoder. In particular, given an input image $\bm{x}^i$, the forward process adds noise to the encoded latent embedding $\bm{z}= \bm{\varepsilon}(\bm{x}^i)$, where $\bm{\varepsilon}(\cdot)$ is an encoder and $\bm{z}_t$ is the noisy latent embedding at timestep $t$. The neural network $\mathcal{D}_{\theta}(\cdot)$ is optimized to predict the presented noise based on image and text instruction conditioning inputs. Formally, the latent diffusion objective is expressed as follows:
\begin{equation}
  \mathcal{L}=\mathbb{E}_{\bm{\varepsilon}(\bm{x}^i), \bm{\varepsilon}({\bm{C}_I}), \bm{c}_T,\epsilon \sim \mathcal{N}(0,1),t}\left[||\epsilon -\mathcal{D}_{\theta}(\bm{z}_t, t,  \bm{\varepsilon}({\bm{C}_I}), \bm{c}_T)||_2^2 \right],
\end{equation}
where $\bm{C}_I$ is the conditioned image, $\bm{C}_T$ is the text editing instruction, and $\hat{\epsilon}_t=\mathcal{D}_{\theta}(\bm{z}_t, t,  \bm{\varepsilon}({\bm{C}_I}), \bm{c}_T)$ is the predicted noise at timestep $t$. Once trained, the estimated latent $\hat{\bm{z}}_{t-1}$ can be derived with a noisy input $\bm{z}_t$ and a predicted noise $\hat{\epsilon}_t$ at timestep $t$.

\begin{figure}
  \setlength{\belowcaptionskip}{0pt}
  \setlength{\abovecaptionskip}{5pt}
  \centering
  \includegraphics[width=0.97\linewidth]{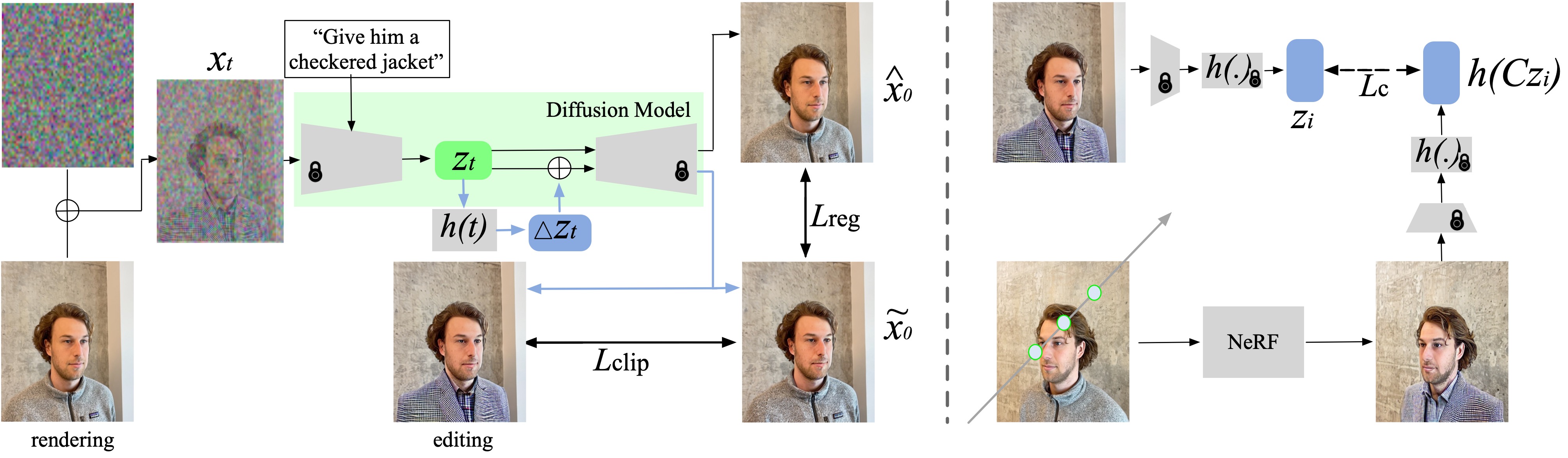}
  \caption{Our pipeline of Edit-DiffNeRF, which is a two-stage framework consisting of a frozen diffusion model, a proposed delta module, and a NeRF. In the first stage, we train the delta module $\bm{h}(t)$ to edit the latent space of a pretrained diffusion model. After training, it is able to produce edited images based on the input text instruction. Then we freeze the weights of the delta module and train the NeRF using those edited images, leveraging the modifications made through the delta module.}
  \label{fig1}
\end{figure}

\section{Method}
Given a pretrained NeRF scene along with the input text instruction, we aim to edit the NeRF scene by manipulating a pretrained and frozen diffusion model as a score function estimator. Which allows us to produce an edited version of the NeRF scene that matches the provided edit instruction. Our pipeline of Edit-DiffNeRF is illustrated in \cref{fig1}, which is a two-stage framework. In the first stage, we train the proposed delta module to edit the latent space of a pretrained diffusion model. Which enables us to make fine-grained modifications to the rendered views and effectively consolidate these instructions in a 3D scene via NeRF training in the second stage.

\subsection{Problem}
In order to supervise the NeRF reconstruction process for editing a 3D scene, many researchers have proposed to utilize a 2D diffusion model for synthesizing the edited outputs \cite{poole2022dreamfusion}. Alternatively, others train a conditional diffusion model by learning $p_{\theta}(\bm{x}_{t-1}|\bm{x}_t, \bm{C}_T)$ \cite{Norman01206}. The experiment results have demonstrated that $\mu_{\theta}(\bm{x}_t,t, \bm{C}_T)$ is closer to the target mean $\widehat{\mu}_{\theta}(\bm{x}_t,t)$ than $\mu_{\theta}(\bm{x}_t,t)$ \cite{ZhangZL22}. This empirical evidence suggests that incorporating conditional inputs can be beneficial to bridge the gap between the predicted posterior mean and the corresponding ground-truth value. However, training diffusion models in image space for each condition can be a cumbersome and impractical task.

\begin{wrapfigure}{r}{10cm}
  \centering
  \includegraphics[width=0.65\textwidth]{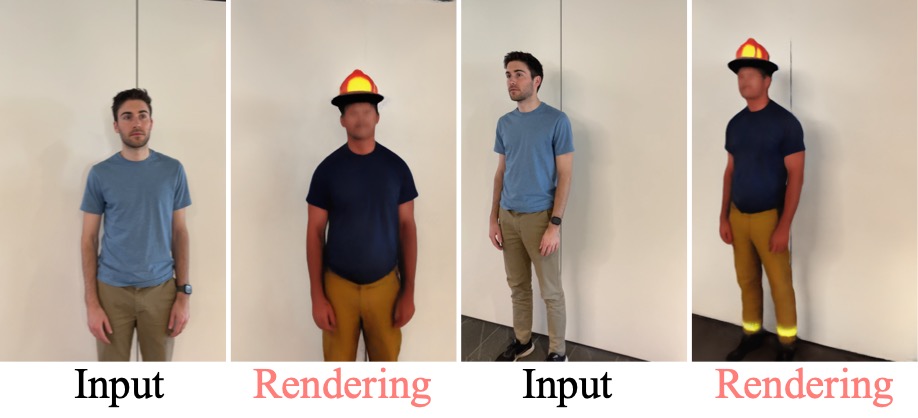}
  \caption{Rendering results with a trained NeRF based on score distillation loss. Edits were performed with a text instruction ``Turn him into a firefighter with a hat''.}
\label{figproblem}
\end{wrapfigure}

Furthermore, an intuitive idea to edit $\bm{x}$ is simply updating $\bm{x}_t$ to optimize NeRF based on the score distillation sampling loss $\nabla \mathcal{L}_{SDS}$ as proposed in DreamFusion \cite{poole2022dreamfusion}. However, it results in a 3D scene with more artifacts or distorted views (see \cref{figproblem}). We believe this is due to a pretrained diffusion model with the fixed semantic latent space may not achieve feasible results in realistic scenarios \cite{Shi12536}.

\subsection{Editing semantic latent space in diffusion models}
In order to manipulate the latent semantic space with the input text instruction and circumvent the process of training an entire diffusion model, we propose to utilize a delta module $\bm{h}(t)$, which learns the shifted latent semantic space $\Delta \bm{z}_t$ given a frozen and pretrained diffusion model and offers significantly reduced computational demands. $\bm{h}(t)$ is implemented as a small compact neural network with two convolutional layers. Which have the same number of channels as the bottleneck layer of the U-Net in the diffusion model. Formally, the parameterization for $\mu_{\theta}(\bm{x}_t,t)$ given the delta module $h(\cdot)$ becomes:
\begin{equation}
  \mu_{\theta}(\bm{x}_t,t, \Delta \bm{z}_t) = \frac{1}{\sqrt{\alpha_t}}\biggl(\bm{x}_t-\frac{\beta_t}{\sqrt{1-\overline{\alpha}_t}}\mathcal{D}_{\theta}(\bm{z}_t|\Delta \bm{z}_t, t, \bm{\varepsilon}({\bm{C}_I}), \bm{c}_T) \biggr).
\end{equation}
Essentially, by introducing the information of $\bm{c}_T$ to the latent semantic space with $\Delta \bm{z}_t$, the predicted $\hat{\epsilon}=\mathcal{D}_{\theta}(\bm{z}_t|\Delta \bm{z}_t, t, \bm{\varepsilon}({\bm{C}_I}), \bm{c}_T)$ is modified. Which, in turn, produces the shifted mean value $\mu_{\theta}(\bm{x}_t,t, \Delta \bm{z}_t)$ to bridge the gap and facilitate the reverse process of reconstructing the provided instructional information in the samples. 

We utilize the architecture of CLIP \cite{RadfordKHRGASAM21}, which comprises an image encoder $\bm{\varepsilon}_I$ and a text encoder $\bm{\varepsilon}_T$ that project inputs to a shared latent space. Building on this, we propose a cross-modal CLIP distance function to evalute the cosine similarity between the input text instruction and the edited image:
\begin{equation}
\mathcal{L}_{\rm clip}=1- \left \langle \bm{\varepsilon}_T(\bm{t}_{\rm src}) -  \bm{\varepsilon}_I(\bm{t}_{\rm tgt}), \bm{\varepsilon}_I(\bm{x}_{\rm src}) - \bm{\varepsilon}_I(\bm{x}_{\rm tgt}) \right \rangle,
\end{equation}
where the input image and its text description are represented by $\bm{x}_{\rm src}$ and $\bm{t}_{\rm src}$. The text instruction and the edited image are denoted as $\bm{t}_{tgt}$ and $\bm{x}_{tgt}$, respectively, and $\left \langle \cdot \right \rangle$ is the cosine similarity operator.
In addition, we add an L1 loss to regulate the produced $\bm{x}_0$ from the original input and the edited one by:
\begin{equation}
  \mathcal{L}_{\rm reg} = \lambda_{\rm reg} |\hat{\bm{x}_0} - \widetilde{\bm{x}}_0|,
\end{equation} 
where $\hat{\bm{x}_0}$ is obtained via the frozen diffusion prior, $\widetilde{\bm{x}}_0$ is generated with the modified latent embedding, and $\lambda_{\rm reg}$ is the hyper-parameter.

\subsection{View-consistent rendering}
Another fundamental challenge when it comes to editing a 3D scene with a 2D diffusion model is that the diffusion models lack 3D awareness, which leads to a generated NeRF with inconsistent and distorted views. Seo et al. \cite{seo2023let} tried to close this gap by utilizing viewpoint-specific depth maps from a coarse 3D structure. However, this embedding needs to optimize a 3D model for each text prompt. This is a computationally intensive process that can still produce blurred and distorted images despite optimization.

To account for the inconsistency challenge, we propose to encode the latent semantic embedding for each view using the pretrained diffusion model and our proposed delta function $\bm{h}(\cdot)$. Specifically, given a pretrained diffusion model and the optimized $\bm{h}(\cdot)$, each rendered image is encoded into a latent embedding $\bm{z}_i$. Inspired by conditional NeRFs \cite{WangCH0022}, the color $\bm{c}$ in NeRF \cite{MildenhallSTBRN20} is extended to a latent-dependent emitted color $\bm{c}_{\bm{z}}$:
\begin{align}
  (\sigma, \bm{c}_{\bm{z}_i}) = F_{\theta}(\bm{x}, \bm{d}, \bm{z}_i),
  \label{eq11}
\end{align}
where the embedding $\bm{z}_i$ is a conditional input, and $F_{\theta}$ is the NeRF model.

\subsection{Multi-view semantic consistency loss}
To achieve our goal, the optimized delta module $\bm{h}(\cdot)$ is supposed to produce consistent latent semantic embeddings as much as possible across different camera poses. Therefore, we propose a novel multi-view semantic consistency loss, denoted as $\mathcal{L}_c$. Which involves using a latent embedding $\bm{z}_i$ extracted from image $\bm{x}^i$ as the conditional input to reconstruct images from different views. The formal expression of our proposed loss $\mathcal{L}_c$ is given as follows:
\begin{equation}
  \mathcal{L}_c = ||\bm{h}(\bm{C}_{\bm{z}_i}) - \bm{z}_i||_1,
\end{equation}
where $\bm{C}_{\bm{z}_i}$ is the rendered output based on \cref{eq11}. Thus the total loss for training a NeRF becomes:
\begin{equation}
  \mathcal{L} = \mathcal{L}_{\rm photo} + \lambda_c \mathcal{L}_c,
\end{equation}
where $\lambda_c$ is the hyper-parameter. By leveraging this strategy, our method ensures consistency across different views while also preserving the integrity of the underlying latent embeddings.

\section{Experiments}
In this section, we first introduce the datasets and implementation details in \cref{sec5.1}. Subsequently, we evaluate the performance of our approach against other methods in \cref{sec5.2}. Finally, to further understand the impact of the key designs, we conduct ablation studies in \cref{sec5.3}.

\begin{figure}[htbp]
  \vspace{-20pt}
	\centering
	\subfloat[Results with larger training datasets]{\includegraphics[width=.48\linewidth]{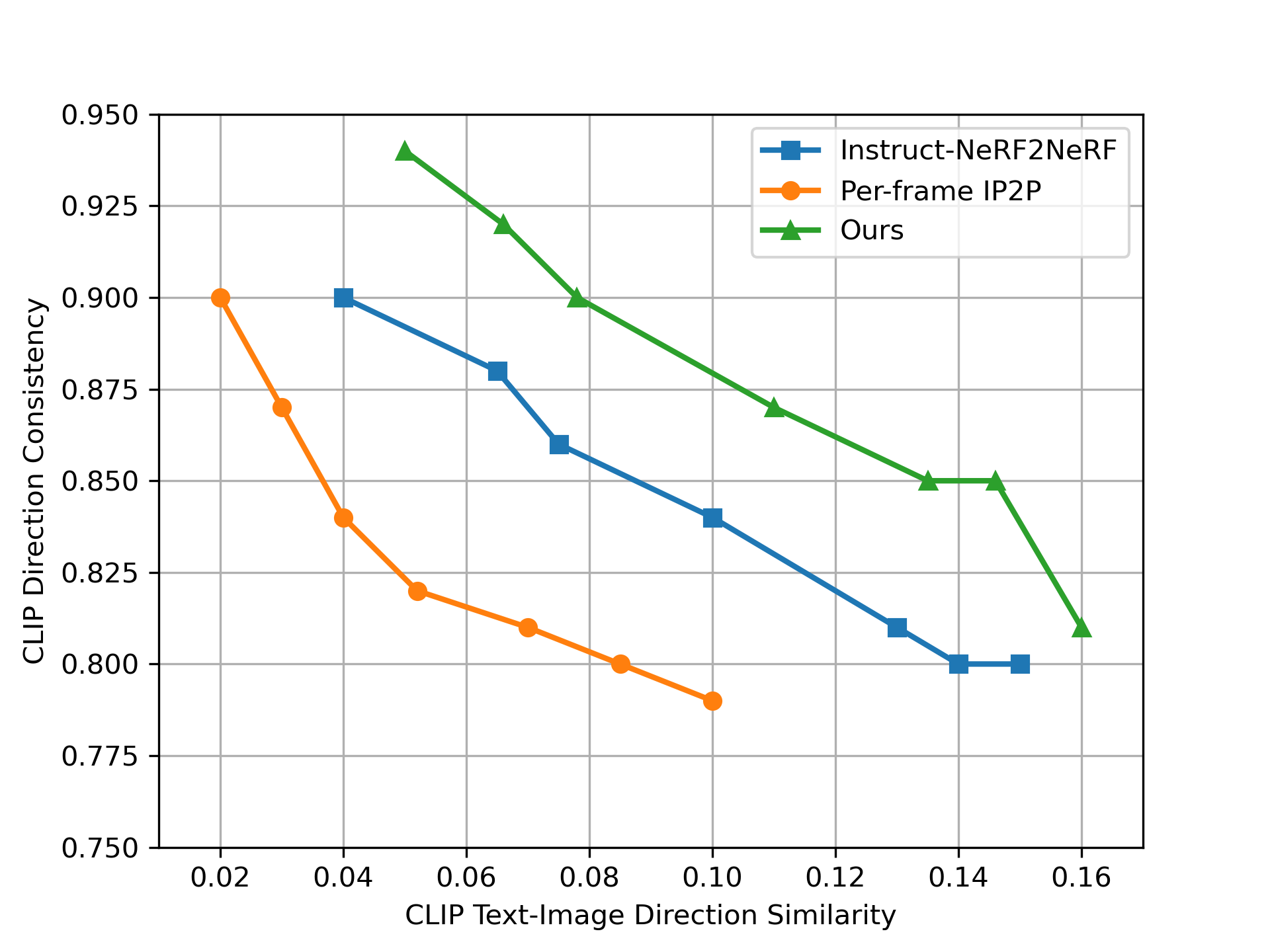}}\hspace{-1pt}
	\subfloat[Results with smaller training datasets (less than 100 images)]{\includegraphics[width=.48\linewidth]{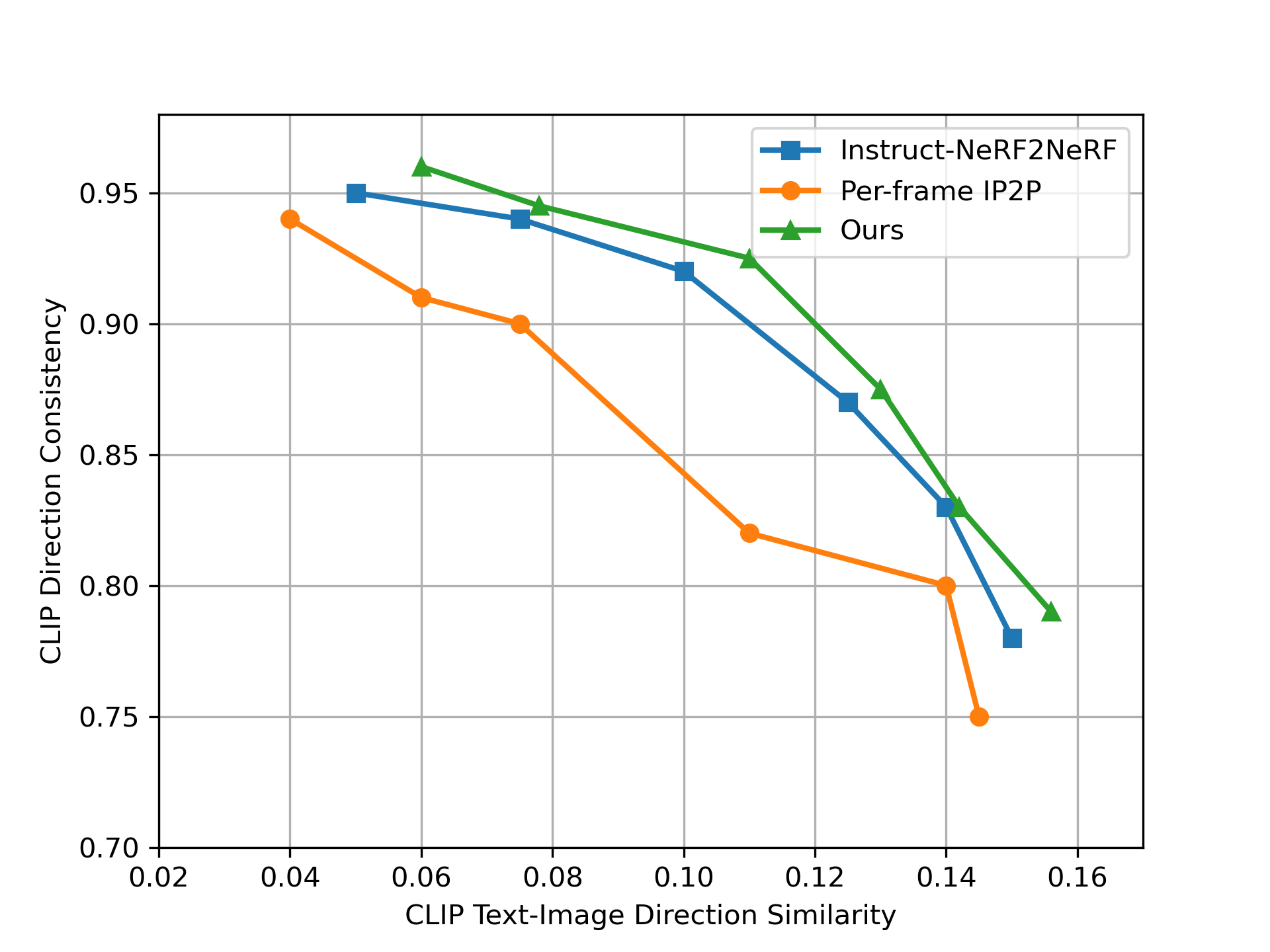}}
	\caption{We plot the trade-off between the CLIP Direction Consistency and the CLIP Text-Image Direction Similarity. For both metrics, higher is better.}
  \label{fig2}
\end{figure}

\begin{table}[htbp]
  \caption{Quantiative evaluation on real-captured scenes}
  \centering
  \begin{tabular}{l|c|c}
    \toprule
    & \makecell[c]{CLIP Text-Image \\ Direction Similarity} & \makecell[c]{CLIP Direction \\ Consistency}  \\
    \midrule
    Per-frame IP2P \cite{Brooks09800} & 0.1603 & 0.8185 \\
    SDS w/ IP2P \cite{Brooks09800} & 0.0266 & 0.9160   \\
    One-time DU \cite{haque2023instructnerf2nerf} & 0.1157 & 0.8823  \\
    Instruct-NeRF2NeRF \cite{haque2023instructnerf2nerf} & 0.1600 & 0.9191 \\
    Ours & \bf{0.2031} & \bf{0.9376} \\
    \bottomrule
  \end{tabular}
  \label{tab1}
\end{table}

\begin{wraptable}{r}{8cm}
  \caption{FID scores}
  \centering
  \begin{tabular}{l|cc|cc}
    \toprule
    & \multicolumn{2}{c|}{Chairs} & \multicolumn{2}{c}{Cars} \\
    \cmidrule(r){1-5}
    & Before & After & Before & After \\
    \midrule
    EditNeRF \cite{LiuZZ0ZR21} & 36.8 & 40.2 & 102.8 & 118.7 \\
    CLIP-NeRF \cite{WangCH0022} & 47.8 & 48.4 & 66.7 & 67.8 \\
    \midrule
    Ours & \bf{32.1} & \bf{32.5} & \bf{48.6} & \bf{49.0} \\
    \bottomrule
  \end{tabular}
  \label{tab11}
\end{wraptable}

\subsection{Experimental setup}
\label{sec5.1}
\textbf{Datasets.} 
We conduct 3D editing on a set of scenes with varying degrees of complexity, including 360-degree scenes of environments and objects, faces, and full-body portraits that are released by \cite{haque2023instructnerf2nerf}. These scenes were captured using two types of cameras: a smartphone and a mirrorless camera. We use the camera poses that were extracted via the COLMAP \cite{SchonbergerF16}. Following CLIP-NeRF \cite{WangCH0022}, we also evaluate the effectiveness of our approach on two publicly available datasets: Photoshape \cite{ParkRFS18} with a collection of 150K chairs and Carla \cite{DosovitskiyRCLK17, Schwarz2020NEURIPS} consisting of 10K cars.

\textbf{Implementation details.}  We choose the official InstructPix2Pix \cite{Brooks09800} as our diffusion prior, which contains a large-scale text-to-image latent diffusion model StableDiffusion \cite{RombachBLEO22}. During the training stage, we uniformly sample timesteps ranging from $t = 1$ to $T = 1000$ for all experiments. The variances of the diffusion process are linearly increased, starting from $\beta_1 = 0.00085$ and reaching $\beta_1 = 0.012$. We optimize our proposed delta module $\bm{h}(t)$ using 50 steps for each view. The training process requires approximately 15 minutes and is performed on four RTX 3090 GPUs. After training, we use the edited images as the supervision to train our NeRF. As the underlying NeRF implementation, we use the nerfacto model from NeRFStudio \cite{nerfstudio}, which is a recommended real-time model tuned for real captures. We follow the training strategy in NeRFStudio and the NeRFs are optimized for 30000 steps with L1 and directional CLIP losses in \cite{RadfordKHRGASAM21}.

\textbf{Metrics.} It should be noted that unlike dynamic NeRF methods, acquiring ground truth views for view synthesis results after editing poses significant challenges, particularly when performing with real scenes. This is primarily because the edited views as a product of user manipulation do not physically exist. Following the evaluation metrics employed in Instruct-NeRF2NeRF \cite{haque2023instructnerf2nerf}, we evaluate two crucial quantitative metrics, namely (1) CLIP Text-Image Direction Similarity, i.e., the alignment between the edited 3D views and the corresponding text instruction, and (2) CLIP Direction Consistency, the temporal consistency of the edit across multiple views \cite{Brooks09800}. Besides, we compute the Fréchet Inception Distance (FID) scores \cite{heusel2018gans} for 2000 rendered images before and after the editing process, which allows us to quantitatively assess the quality and fidelity of the edited scenes.

\begin{figure}[htbp]
  \subfloat{
    \begin{minipage}[h]{0.24\textwidth}
\includegraphics[width=0.95\linewidth, height=0.75\linewidth]{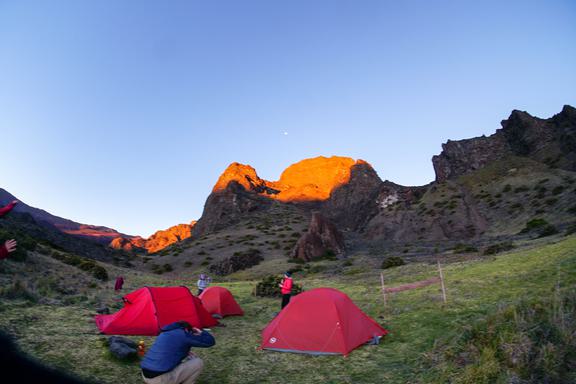}
 \includegraphics[width=0.95\linewidth, height=0.75\linewidth]{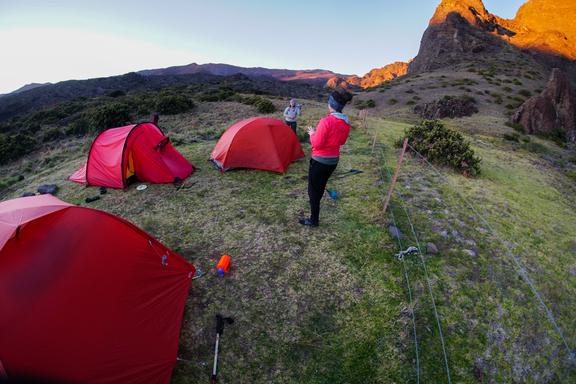}
 \includegraphics[width=0.95\linewidth, height=0.75\linewidth]{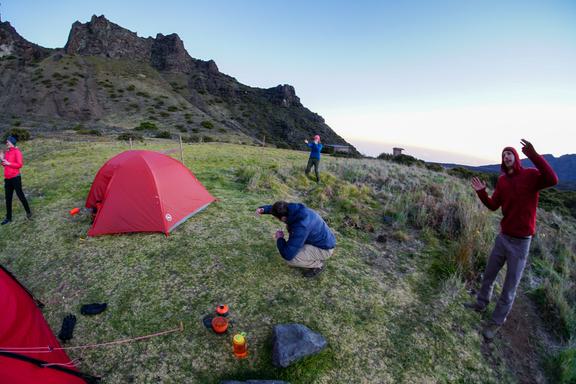}
 \includegraphics[width=0.95\linewidth, height=0.75\linewidth]{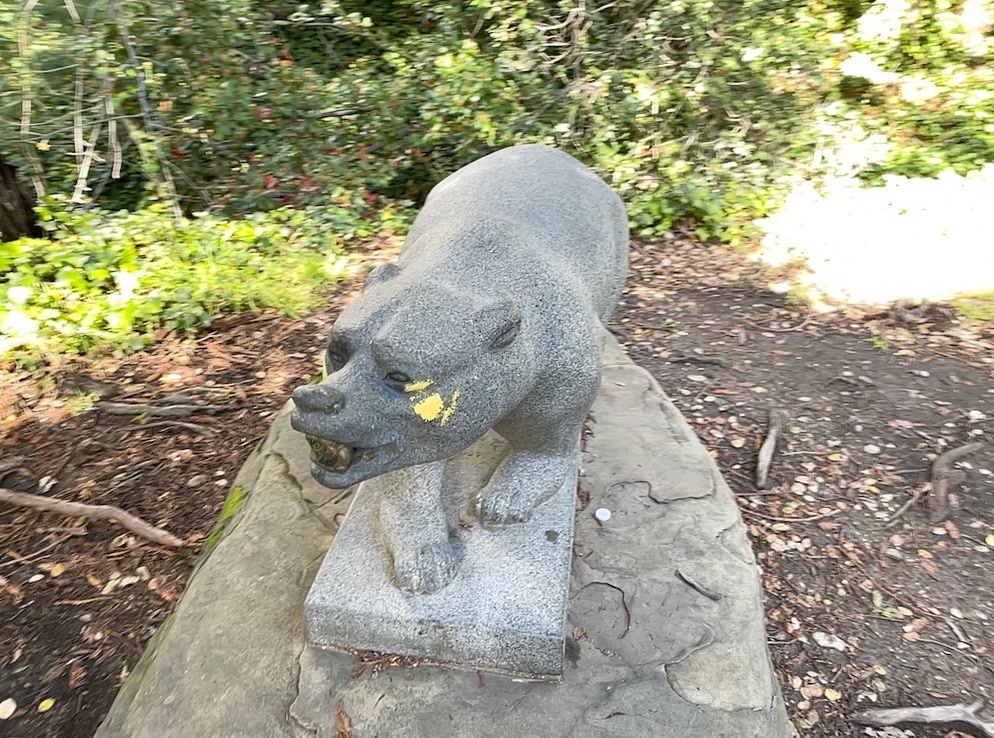}
 \includegraphics[width=0.95\linewidth, height=0.75\linewidth]{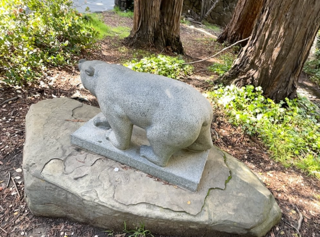}
 \includegraphics[width=0.95\linewidth, height=0.75\linewidth]{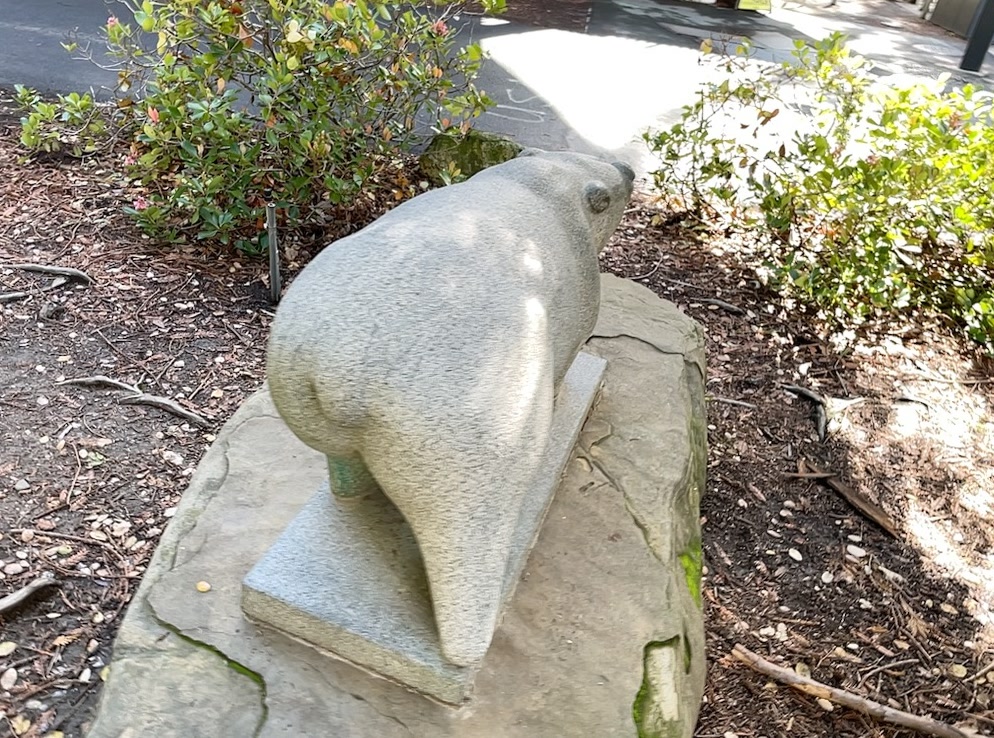}
 \centerline{Original NeRF}
\label{fig:short-4a}
    \end{minipage}}
    \hspace{-7pt}
\subfloat{
        \begin{minipage}[h]{0.24\textwidth}
    \includegraphics[width=0.98\linewidth, height=0.75\linewidth]{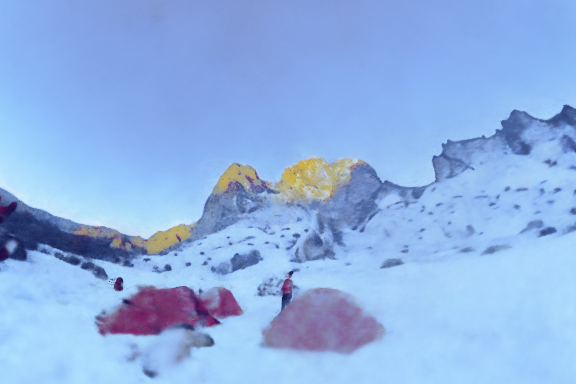}
     \includegraphics[width=0.98\linewidth, height=0.75\linewidth]{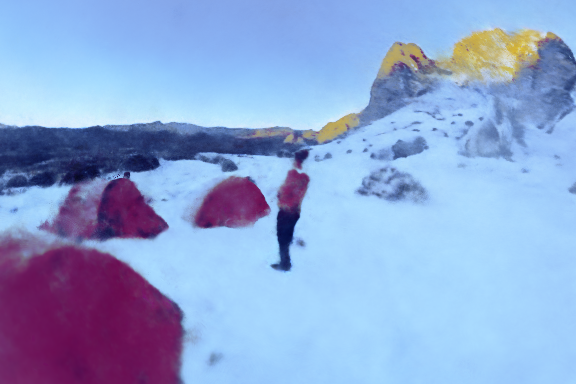}
     \includegraphics[width=0.98\linewidth, height=0.75\linewidth]{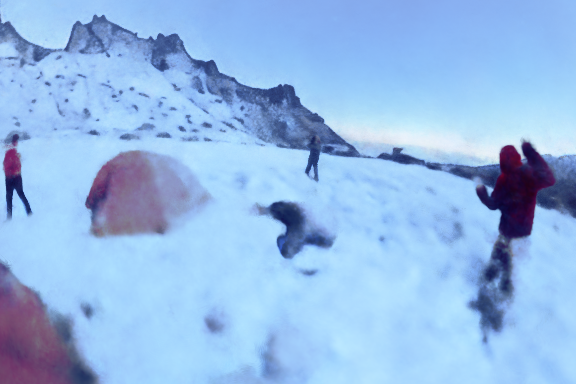}
     \includegraphics[width=0.98\linewidth, height=0.75\linewidth]{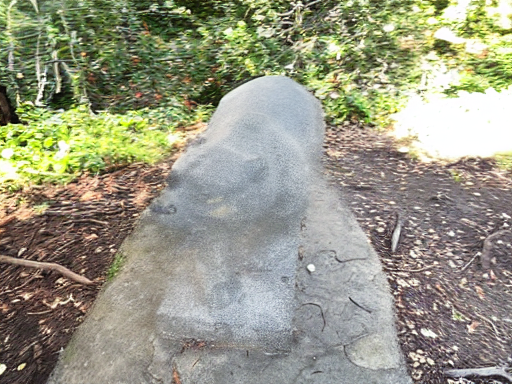}
     \includegraphics[width=0.98\linewidth, height=0.75\linewidth]{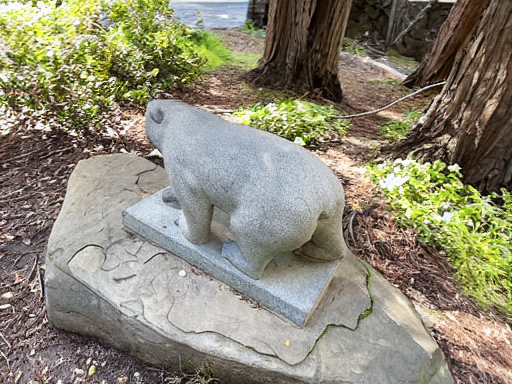}
     \includegraphics[width=0.98\linewidth, height=0.75\linewidth]{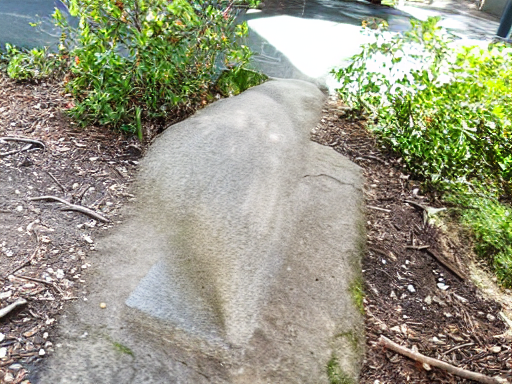}
     \centerline{InstructPix2Pix \cite{Brooks09800}}
    \label{fig:short-3a}
        \end{minipage}}
        \hspace{-5pt}
\subfloat{
        \begin{minipage}[h]{0.24\textwidth}
    \includegraphics[width=0.98\linewidth, height=0.75\linewidth]{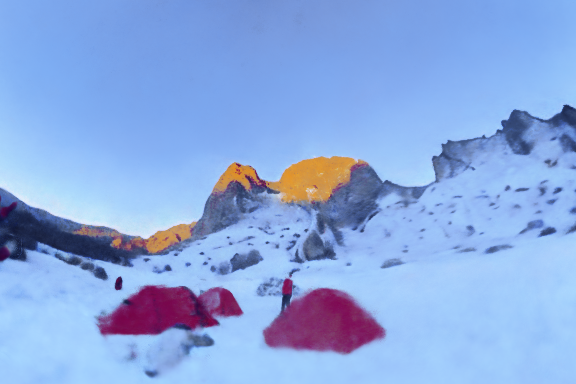}
     \includegraphics[width=0.98\linewidth, height=0.75\linewidth]{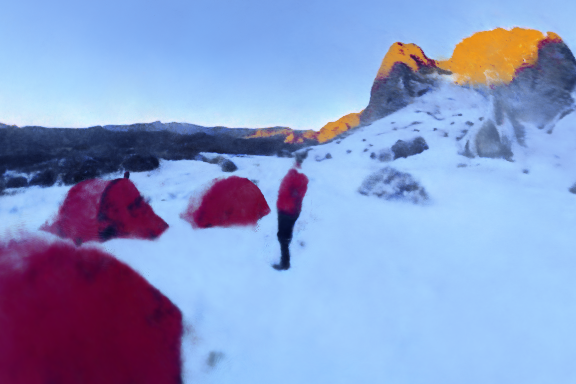}
     \includegraphics[width=0.98\linewidth, height=0.75\linewidth]{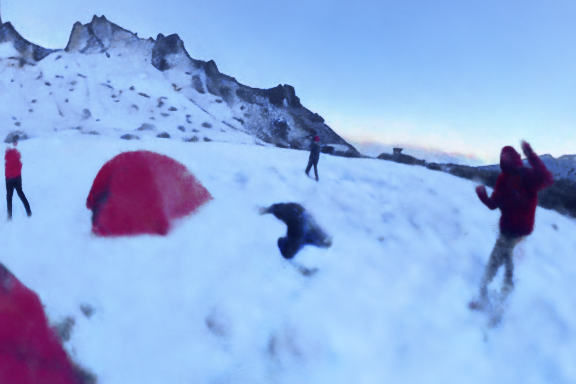}
     \includegraphics[width=0.98\linewidth, height=0.75\linewidth]{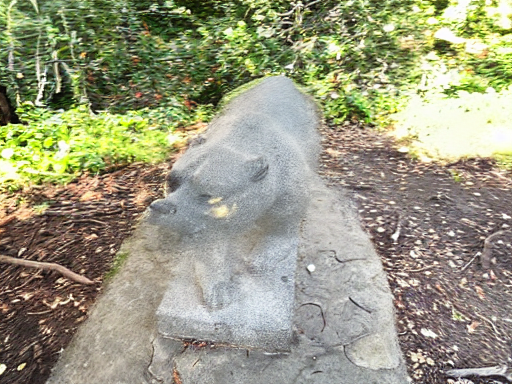}
     \includegraphics[width=0.98\linewidth, height=0.75\linewidth]{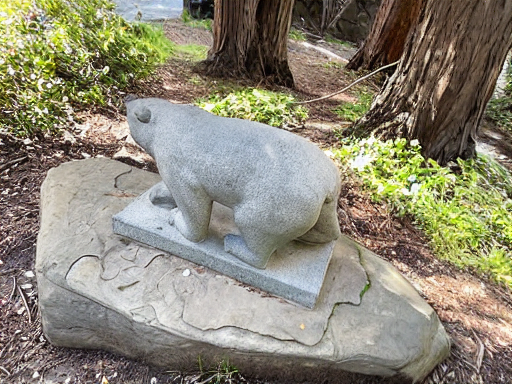}
     \includegraphics[width=0.98\linewidth, height=0.75\linewidth]{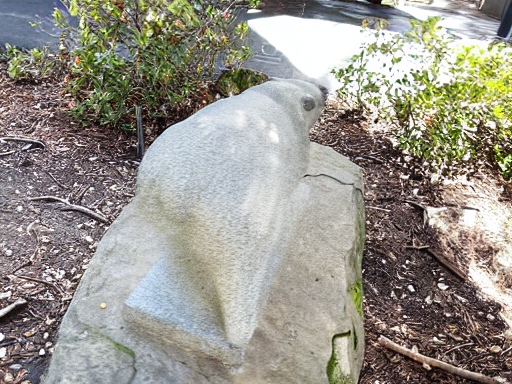}
     \centerline{Instruct-NeRF2NeRF \cite{haque2023instructnerf2nerf}}
    \label{fig:short-3b}
        \end{minipage}}
        \hspace{-5pt}
\subfloat{
        \begin{minipage}[h]{0.24\textwidth}
     \includegraphics[width=0.98\textwidth,height=0.75\textwidth]{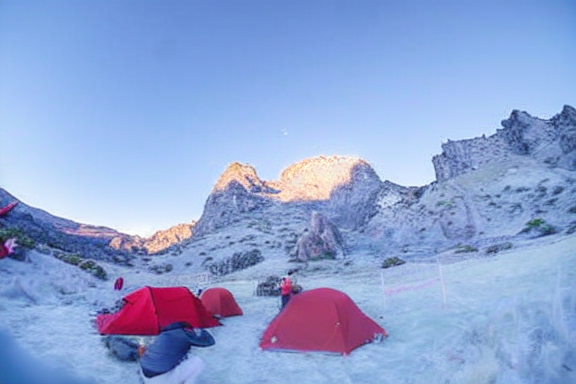}
     \includegraphics[width=0.98\linewidth, height=0.75\linewidth]{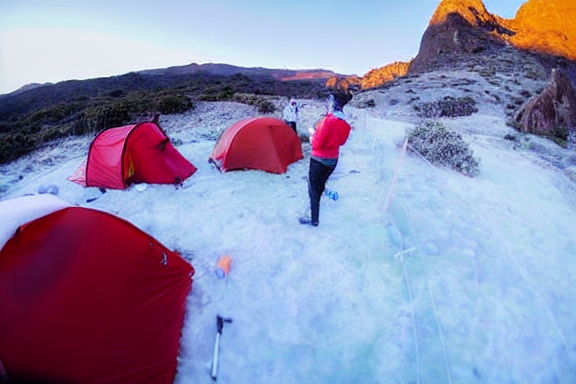}
     \includegraphics[width=0.98\linewidth, height=0.75\linewidth]{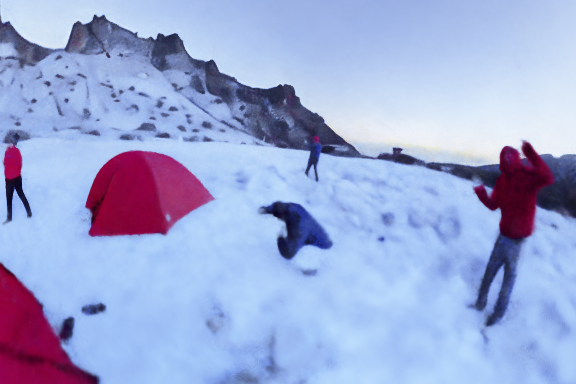}
     \includegraphics[width=0.98\linewidth, height=0.75\linewidth]{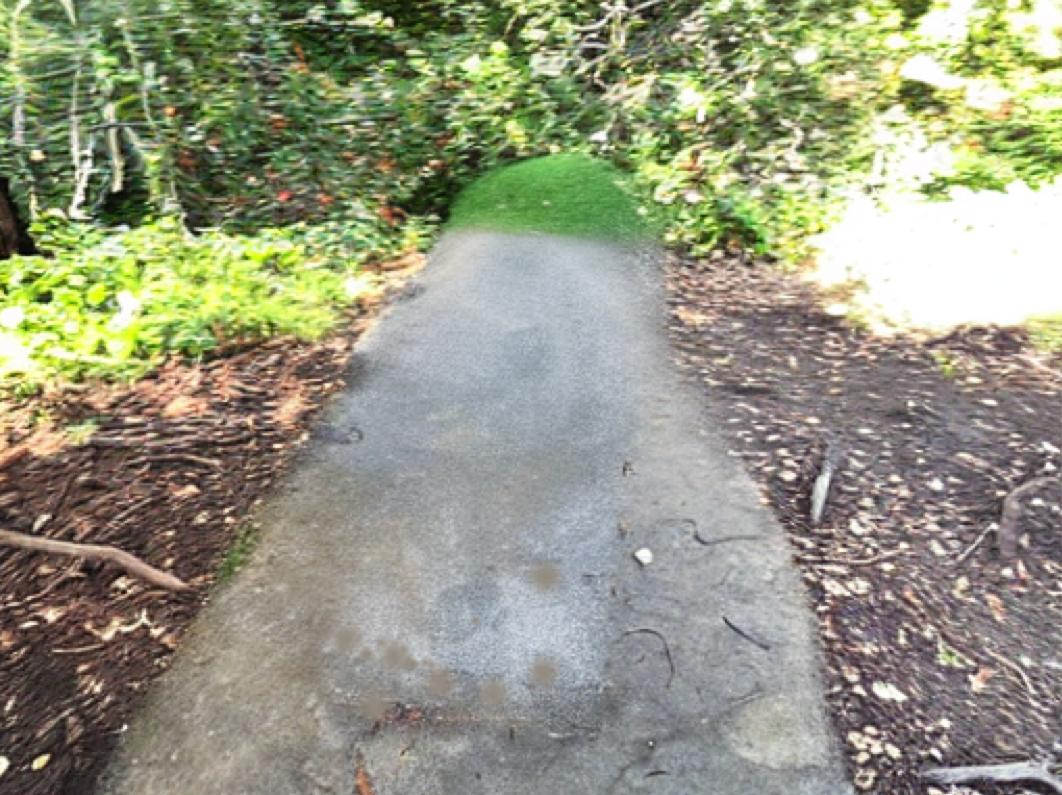}
     \includegraphics[width=0.98\linewidth, height=0.75\linewidth]{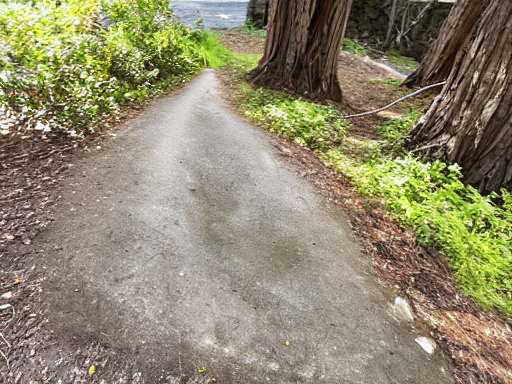}
     \includegraphics[width=0.98\linewidth, height=0.75\linewidth]{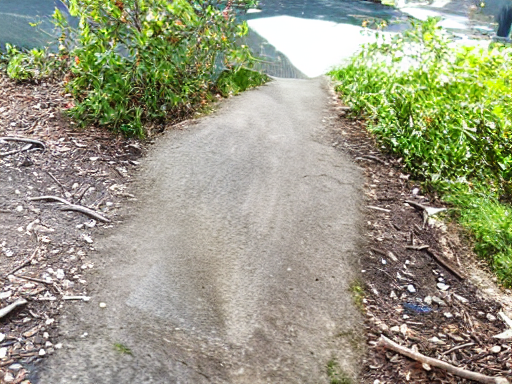}
     \centerline{\bf{Ours}}
    \label{fig:short-3c}
        \end{minipage}}
        \subfloat{
        \begin{minipage}[h]{0.03\textwidth}
          \rotatebox{90}{~~~~~~~~~~~~~~~~~~~~~~~~~~~~\scalebox{1.0}[1.0]{``Make it look like just snowed''}}
     \rotatebox{90}{\scalebox{1.0}[1.0]{``Delete the bear statue and the stage''}}
     \rotatebox{90}{~~~~~~~~~~~~~~~~~\scalebox{1.0}[1.0]{}~~~~~~~~~}
        \end{minipage}}
        \hspace{-6pt}
\caption{Visual comparisons with a collection of recent state-of-the-art methods.}
\vspace{-10pt}
\label{fig3}
\end{figure}

\subsection{Quantiative evaluation}
\label{sec5.2}
\cref{tab1} and \cref{tab11} show the superiority of Edit-DiffNeRF over other state-of-the-art methods on real scenes. We observe that by learning the optimal delta module to edit the latent semantic space, our Edit-DiffNeRF can have notably higher CLIP Text-Image Direction Similarity and Consistency. In \cref{fig2}, we also plot the trade-off between the CLIP Direction Consistency and the CLIP Text-Image Direction Similarity over two scenes. As these two metrics compete with each other, when the degree to which the output images align with the desired edit increases, the consistency with the input image decreases. As is observed from \cref{fig2}, compared to the recent state-of-the-art method Instruct-NeRF2NeRF \cite{haque2023instructnerf2nerf}, the CLIP Direction consistency obtained by our Edit-DiffNeRF is significantly higher, even with similar CLIP Text-Image Direction Similarity values. Furthermore, we observe that when the training dataset for a scene is smaller (consisting of less than 100 images), both the Instruct-NeRF2NeRF \cite{haque2023instructnerf2nerf} and our method yield similar results, with lower directional similarity.

In \cref{tab11}, we report the FID scores for measuring the image quality of synthesized views before and after editing. To calculate the FID scores for rendered images, we employ a set of 2000 randomly selected test images. Subsequently, we apply various edit instructions similar to CLIP-NeRF \cite{WangCH0022} to these images and recompute the FID scores for the edited results. On the chair dataset, When evaluated on the chair dataset, EditNeRF \cite{LiuZZ0ZR21} demonstrates improved performance in terms of reconstruction compared to CLIP-NeRF \cite{WangCH0022}. However, it is worth noting that the quality of the edited images noticeably decreases after the editing process. When evaluated on the car dataset, CLIP-NeRF \cite{WangCH0022} exhibits a significant improvement over EditNeRF \cite{LiuZZ0ZR21} in terms of reconstruction quality before and after editing. Finally, compared to those two methods, our Edit-DiffNeRF not only greatly improves the overall quality but also effectively preserves the image quality after the editing process.  

\begin{wrapfigure}{r}{10cm}
  \centering
  \includegraphics[width=0.65\textwidth]{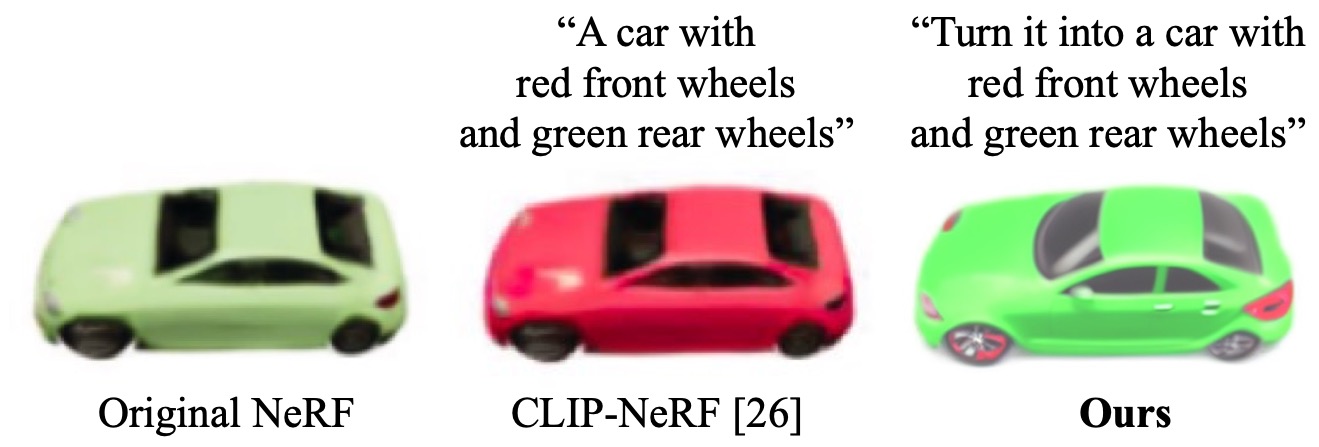}
  \caption{Comparisons of editing results between CLIP-NeRF \cite{WangCH0022} and our Edit-DiffNeRF.}
\label{fig4}
\end{wrapfigure}

\textbf{Editing results.}
We show edited results rendered from different views in \cref{fig3} for real-captured scenes. For comparison, we also show the original NeRF rendering results under the same views before editing. In \cref{fig3}, the first set is a campsite scene. We edited it by a text instruction ``Make it look like just snowed''. On the one hand, as is observed from \cref{fig3}, the generated results via the InstructPix2Pix \cite{Brooks09800} are with ambiguity on what exactly to edit and exhibit considerable inconsistencies across different views. On the other hand, although the Instruct-NeRF2NeRF \cite{haque2023instructnerf2nerf} appears to produce feasible views, some of them tend to show significant variance, resulting in a 3D scene that is blurry and highly distorted. Instead, the rendered multiple views obtained via our Edit-DiffNeRF show its capability to effectively edit real-world scenes, surpassing the achievements of previous approaches by delivering photo-realistic results while maintaining 3D consistency. In the second set of \cref{fig3}, the images were edited by instruction "Delete the bear statue and the stage". According to the figure, neither InstructPix2Pix \cite{Brooks09800} nor Instruct-NeRF2NeRF \cite{haque2023instructnerf2nerf} are able to obtain the intended editing results effectively. This limitation primarily arises from the incapacity of InstructPix2Pix to handle large spatial manipulations.

\begin{wrapfigure}{r}{10cm}
  \vspace{-25pt}
  \centering
  \includegraphics[width=0.65\textwidth]{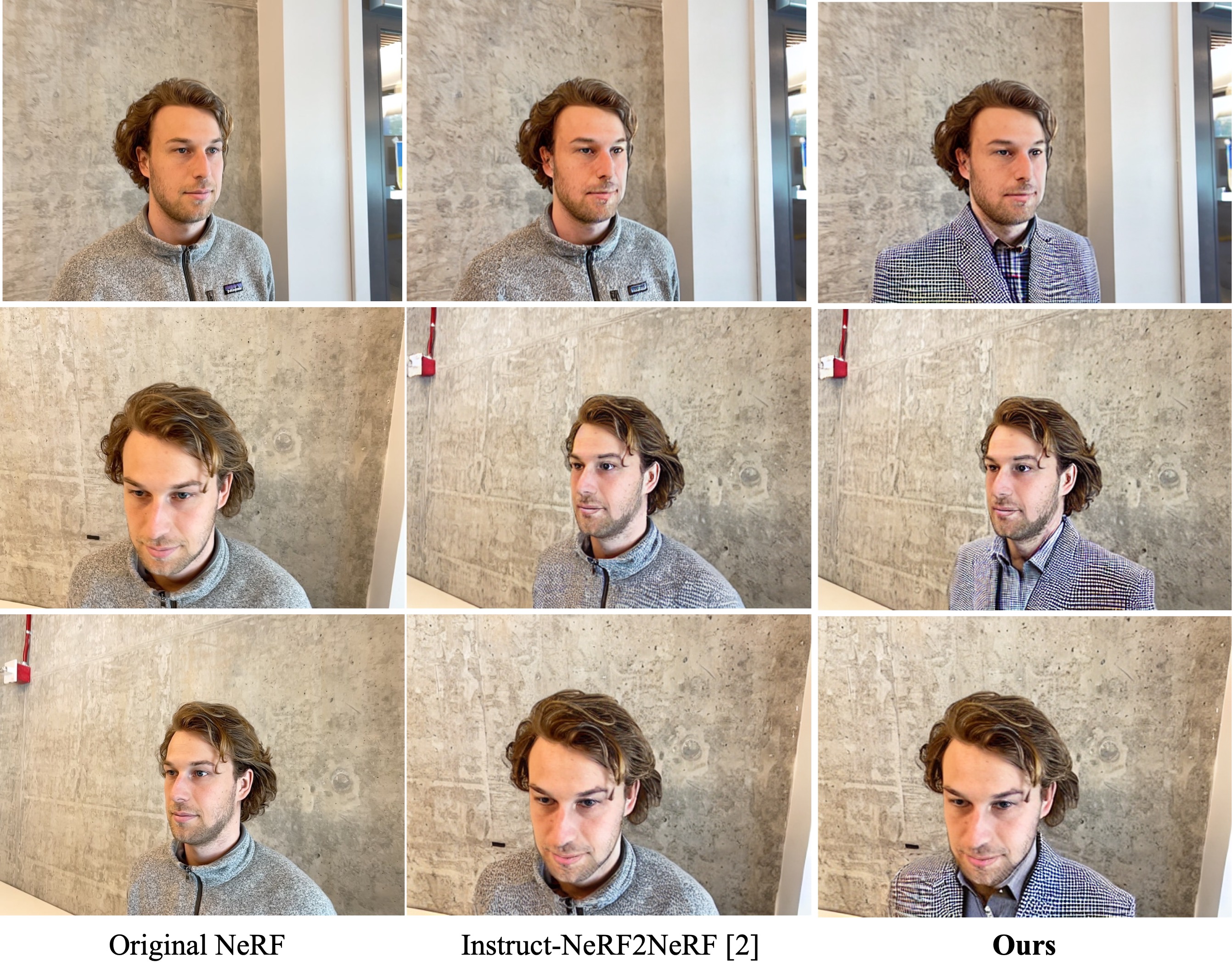}
\caption{Comparison with Instruct-NeRF2NeRF \cite{haque2023instructnerf2nerf}. Edits were performed with a text instruction ``Give him a checkered jacket''.}
\vspace{-30pt}
\label{fig5}
\end{wrapfigure}

Moreover, we present additional experimental results along with NeRF rendering results in \cref{fig4} and \cref{fig5} to further illustrate our findings. The comparisons in \cref{fig4} were performed with a text instruction ``Turn it into a car with red front wheels and green rear wheels''. It can be noted that CLIP-NeRF \cite{WangCH0022} is not able to handle fine-grained edits. Instead, our edited result (right column) is  quite consistent with the instruction. In \cref{fig5}, we can further observe the presence of inconsistent edits within Instruct-NeRF2NeRF \cite{haque2023instructnerf2nerf} that fail to consolidate in a 3D scene (middle column). Instead, our model provides a superior solution (right column), rendering consistent results and allows for significant visual manipulations.

\subsection{Ablation study}
\label{sec5.3}
We validate the effectiveness of our Edit-DiffNeRF by conducting a comprehensive comparative analysis between our approach and several other variants.

\begin{wraptable}{r}{10cm}
  \caption{Ablation study results}
  \centering
  \begin{tabular}{l|c|c}
    \toprule
    & \makecell[c]{CLIP Text-Image \\ Direction Similarity} & \makecell[c]{CLIP Direction \\ Consistency}  \\
    \midrule
    Instruct-NeRF2NeRF \cite{haque2023instructnerf2nerf} & 0.1120 & 0.7805 \\
    Ours w/o $\mathcal{L}_c$ & 0.1703 & 0.9198 \\
    Ours & \bf{0.2031} & \bf{0.9376} \\
    \bottomrule
  \end{tabular}
  \label{tab2}
\end{wraptable}

\textbf{Impact of delta module.}
We first compare our model with against Instruct-NeRF2NeRF \cite{haque2023instructnerf2nerf}. We use the official code released by \cite{Brooks09800} and fine-tune the entire UNet model in InstructPix2Pix \cite{Brooks09800} to edit real images. \cref{tab2} demonstrates that our proposed Edit-DiffNeRF outperforms InstructPix2Pix in all aspects. We attribute this superiority to the fact that fine-tuning the entire model for each scene is challenging, thereby resulting in inferior results.

\textbf{Impact of multi-view semantic consistency loss.}
In addition, we perform experiments where we exclude the consistency loss $\mathcal{L}_c$. As is observed from \cref{tab2}, even in the absence of the multi-view semantic consistency loss $\mathcal{L}_c$, our method still surpasses Instruct-NeRF2NeRF \cite{haque2023instructnerf2nerf} with a trained delta module. Nevertheless, without this loss, our method still lacks 3D consistency and only achieves a slight performance gain. In contrast, incorporating this loss yields substantial improvements in the results.

\section{Conclusion}
In this paper, we have outlined the underlying challenges in achieving accurate NeRF scene modifications with pretrained 2D diffusion models. To address this limitation, we introduce the Edit-DiffNeRF framework, which specifically targets editing the semantic latent space within pretrained diffusion models. Specifically, the Edit-DiffNeRF framework is devised to learn latent semantic directions using a delta module, guided by provided text instructions, which allows for the effective consolidation of these instructions within a 3D scene through NeRF training. Furthermore, we introduce a multi-view semantic consistency loss to ensure semantic consistency across different views. Extensive experiments demonstrate that our approach consistently and effectively enables edits across a wide range of real-captured scenes. Moreover, it significantly improves the text-image consistency of the edited results. 

\section{Limitations}
Despite the encouraging performance gained by our work, it suffers from two limitations. First, our model is subject to the visual quality of the rendered images generated using NeRF techniques, as well as the diffusion model's ability to generalize to arbitrary edits. Second, the quality of edited images decreases when the input images have low resolution or are out of focus.

{\small
\bibliographystyle{ieee_fullname}
\bibliography{mybib}
}


\end{document}